\documentclass[11pt,a4paper]{article}
\pdfoutput=1
\usepackage[hyperref]{acl2017}
\usepackage{times}
\usepackage{latexsym}
\usepackage{url}
\usepackage{amsmath,graphicx}
\usepackage{amsfonts}
\usepackage{amssymb}
\usepackage{multirow}
\usepackage{algorithm}
\usepackage{algorithmic}
\usepackage{graphicx}
\usepackage{booktabs}
\usepackage[bottom]{footmisc}
\usepackage{subfig}

\newcommand{\vect}[1]{\boldsymbol{#1}}
\newcommand\defeq{\mathrel{\overset{\makebox[0pt]{\mbox{\tiny def}}}{=}}}

\begin{document}

\title{Transfer Learning for Neural  Semantic Parsing}

\author{Xing Fan, Emilio Monti, Lambert Mathias, Markus Dreyer}

\maketitle

\begin{abstract}

The goal of semantic parsing is to map natural language to a machine
interpretable meaning representation language (MRL). One of the constraints
that limits full exploration of deep learning technologies for semantic parsing
is the lack of sufficient annotation training data. In this paper, we propose
using sequence-to-sequence in a multi-task setup for semantic parsing with a focus on transfer learning. We explore three multi-task architectures for sequence-to-sequence modeling and compare their performance with an independently trained model. Our experiments show that the multi-task setup aids transfer learning from an auxiliary task with large labeled data to a target task with smaller labeled data. We see absolute accuracy gains ranging from 1.0\% to 4.4\% in our in-house data set, and we also see good gains ranging from 2.5\% to 7.0\% on the ATIS semantic parsing tasks with syntactic and semantic auxiliary tasks.


\end{abstract}

\section{Introduction}

Conversational agents, such as Alexa, Siri and Cortana, solve complex tasks by interacting and mediating between the end-user and multiple backend software applications and services. Natural language is a simple interface used for communication between these agents. However, to make natural language machine-readable we need to map it to a representation that describes the semantics of the task expressed in the language. Semantic parsing is the process of mapping a natural-language sentence into a formal machine-readable representation of its meaning. This poses a challenge in a multi-tenant system that has to interact with multiple backend knowledge sources each with their own semantic formalisms and custom schemas for accessing information, where each formalism has various amount of annotation training data.

Recent works have proven sequence-to-sequence to be an effective model architecture~\cite{jia2016data, dong2016language} for semantic parsing. However, because of the limit amount of annotated data, the advantage of neural networks to capture complex data representation using deep structure~\cite{johnson2016google}  has not been fully explored. Acquiring data is expensive and sometimes infeasible for task-oriented systems, the main reasons being multiple formalisms (e.g., SPARQL for WikiData~\cite{vrandevcic2014wikidata}, MQL for Freebase~\cite{flanagan2008mql}), and multiple tasks (question answering, navigation interactions, transactional interactions). We propose to exploit these multiple representations in a multi-task framework so we can minimize the need for a large labeled corpora across these formalisms. By suitably modifying the learning process, we capture the common structures that are implicit across these formalisms and the tasks they are targeted for.

In this work, we focus on a sequence-to-sequence based transfer learning for semantic parsing. In order to tackle the challenge of multiple formalisms, we apply three multi-task frameworks with different levels of parameter sharing. Our hypothesis is that the encoder-decoder paradigm learns a canonicalized representation across all tasks. Over a strong single-task sequence-to-sequence baseline, our proposed approach shows accuracy improvements across the target formalism. In addition, we show that even when the auxiliary task is syntactic parsing we can achieve good gains in semantic parsing that are comparable to the published state-of-the-art.

\section{Related Work}

There is a large body of work for semantic parsing. These approaches fall into three broad categories -- completely supervised learning based on fully annotated logical forms associated with each sentence~\cite{zelle1996learning, zettlemoyer2012learning}   using question-answer pairs and conversation logs as supervision~\cite{artzi2011bootstrapping, liang2011learning, berant2013semantic} and distant supervision~\cite{cai2013semantic,reddy2014large}. All these approaches make assumptions about the task, features and the target semantic formalism.

On the other hand, neural network based approaches, in particular the use of recurrent neural networks (RNNs) and encoder-decoder paradigms~\cite{sutskever2014sequence}, have made fast progress on achieving state-of-the art performance on various NLP tasks~\cite{vinyals2015grammar, dyer2015transition, bahdanau2014neural}. A key advantage of RNNs in the encoder-decoder paradigm is that very few assumptions are made about the domain, language and the semantic formalism. This implies they can generalize faster with little feature engineering. 

Full semantic graphs can be expensive to annotate, and  efforts to date have been fragmented across different formalisms, leading to a limited amount of annotated data in any single formalism.  Using neural networks to train semantic parsers on limited data is quite challenging. Multi-task learning aims at improving the generalization performance of a task using related tasks~\cite{caruana1998multitask, ando2005framework, smith2004}. This opens the opportunity to utilize large amounts of data for a related task to improve the performance across all tasks. There has been recent work in NLP demonstrating improved performance for machine translation~\cite{dong2015multi} and syntactic parsing~\cite{luong2015multi}.

In this work, we attempt to merge various strands of research using sequence-to-sequence modeling for semantic parsing with focusing on improving semantic formalisms with small amount of training data using a multi-task model architecture. The closest work is \newcite{Jonathan2017multiKB}. Similar to this work, the authors use a neural semantic parsing model in a multi-task framework to jointly learn over multiple knowledge bases.  Our work differs from their work in that we focus our attention on transfer learning, where we have access to a large labeled resource in one task and want another semantic formalism with access to limited training data to benefit from a multi-task learning setup. Furthermore, we also demonstrate that we can improve semantic parsing tasks by using large data sources from an auxiliary task such as syntactic parsing, thereby opening up the opportunity for leveraging much larger datasets. Finally, we carefully compare multiple multi-task architectures in our setup and show that increased sharing of both the encoder and decoder along with shared attention results in the best performance.


\section{Problem Formulation}

\subsection{Sequence-to-Sequence Formulation}

\begin{figure*}[ht]
        \includegraphics[clip, height=9cm, width=12cm]{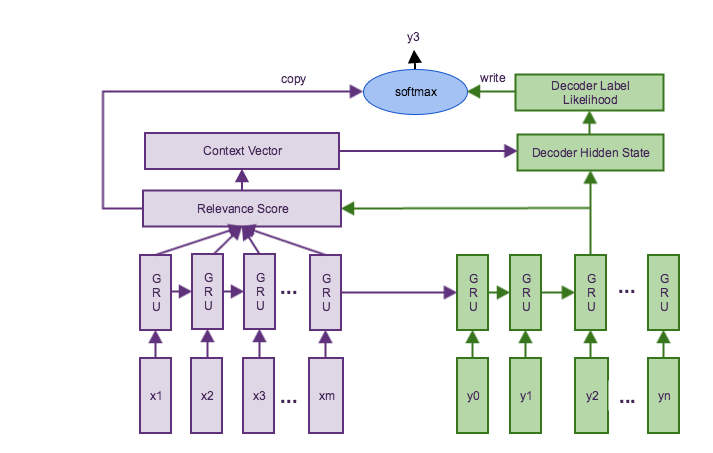}
        \caption{{An example of how the decoder output $y_3$ is generated.}}
        \label{figure:copy}
\end{figure*}

Our semantic parser extends the basic encoder-decoder approach in~\newcite{jia2016data}. Given a sequence of inputs $\vect{x}={x_1,\dots,x_m}$, the sequence-to-sequence model will generate an output sequence of $\vect{y}={y_1,\dots,y_n}$.
We encode the input tokens $\vect{x}={x_1,\dots,x_m}$ into a sequence of embeddings $\vect{h}= \vect{h}_1,\dots,\vect{h}_m$ 
\begin{equation}
\vect{h}_i = f_{\textnormal{encoder}}(E_x(x_i), \vect{h}_{i-1})
\end{equation}
First, an input embedding layer $E_x$ maps each word $x_i$ to a fixed-dimensional vector which is then fed as input to the network $f$ to obtain the hidden state representation $\vect{h}_i$. The embedding layer $E_x$ could contain one single word embedding lookup table or a combination of word and gazetteer embeddings, where we concatenate the output from each table. For the encoder and decoder, we use a stacked Gated Recurrent Units (GRU) ~\cite{cho2014learning}.\footnote{In order to speedup training, we use a right-to-left GRU instead of a bidirectional GRU.} The hidden states are then converted to one fixed-length context vector per output index, $\vect{c}_j=\phi_j(h_1,\dots,h_m)$, where $\phi_j$ summarizes all input hidden states to form the context for a given output index $j$.\footnote{In a vanilla decoder, each $\phi_j(h_1,\dots,h_m)\defeq h_m$, i.e, the hidden representation from the last state of the encoder is used as context for every output time step $j$.}

The decoder then uses these fixed-length vectors $\vect{c}_j$ to create the target sequence through the following model. At each time step $j$ in the output sequence, a state $\vect{s}_j$ is calculated as
\begin{equation}
\vect{s}_j = f_{\textnormal{decoder}}(E_y(y_{j-1}), \vect{s}_{j-1}, \vect{c}_j)
\end{equation}
Here, $E_y$ maps any output symbol to a fixed-dimensional vector. Finally, we compute the probability of the output symbol $y_j$ given the history $y_{< j}$ using Equation~\ref{eq:wo_copy}.

\begin{equation} \label{eq:wo_copy}
p(y_j\mid y_{<j}, \vect{x}) \propto \exp(\vect{O}[\vect{s}_{j}; \vect{c}_{j}])
\end{equation}

where the matrix $\vect{O}$ projects the concatenation of $\vect{s}_{j}$ and $\vect{c}_{j}$, denoted as $[\vect{s}_{j}; \vect{c}_{j}]$, to the final output space. The matrix $\vect{O}$ are part of the trainable model parameters.
We use an attention mechanism~\cite{bahdanau2014neural} to summarize the context vector $\vect{c}_j$,
\begin{equation}
\vect{c}_j=\phi_j(h_1,\dots,h_m) = \sum_{i=1}^{m} \alpha_{ji}\ \vect{h}_i
\end{equation}

where $j\in [1,\dots,n]$ is the step index for the decoder output and $\alpha_{ji}$ is the attention weight, calculated using a softmax:
\begin{equation}
\alpha_{ji} = \frac{\exp(e_{ji})}{\sum_{i'=1}^{m} \exp(e_{ji'})}
\end{equation}

where $e_{ji}$ is the relevance score of each context vector $\vect{c}_j$, modeled as:
\begin{equation}
e_{ji}=g(\vect{h}_i, \vect{s}_{j})
\end{equation}

In this paper, the function $g$ is defined as follows:
\begin{equation}
g(\vect{h}_i, \vect{s}_{j}) = \boldsymbol{\upsilon}^\intercal\tanh(\vect{W}_{1}\vect{h}_i + \vect{W}_{2}\vect{s}_{j})
\end{equation}

where $\boldsymbol{\upsilon}$, $\vect{W}_1$ and $\vect{W}_2$ are trainable parameters.

In order to deal with the large vocabularies in the output layer introduced by the long tail of entities in typical semantic parsing tasks, we use a copy mechanism~\cite{jia2016data}. At each time step $j$, the decoder chooses to either copy a token from the encoder's input stream or to write a token from the the decoder's fixed output vocabulary. We define two actions: 

\begin{enumerate}
 \item $\textsc{Write}[y]$ for some $y \in \mathcal{V}_{\textnormal{decoder}}$, where $\mathcal{V}_{\textnormal{decoder}}$ is the output vocabulary of the decoder.
 \item $\textsc{Copy}[i]$  for some $i \in {1,\dots,m}$, which copies one symbol from the $m$ input tokens.
\end{enumerate}

We formulate a single softmax to select the action to take, rewriting Equation~\ref{eq:wo_copy} as follows:
\begin{align}
p(a_j = \textsc{Write}[y_j]\mid y_{<j}, \vect{x}) &\propto \exp(\vect{O}[\vect{s}_{j};\vect{c}_{j}])\\
p(a_j = \textsc{Copy}[i]\mid y_{<j}, \vect{x}) &\propto \exp(e_{ji})
\end{align}

The decoder is now a softmax over the actions $a_j$; Figure~\ref{figure:copy} shows how the decoder's output $\vect{y}$ at the third time step $y_3$ is generated. At each time step, the decoder will make a decision to copy a particular token from input stream or to write a token from the fixed output label pool.

\begin{figure*}[ht]
\centering
\subfloat[\textit{one-to-many}: A multi-task architecture where only the encoder is shared across the two tasks.]{%
  \includegraphics[clip, width = 110mm, height = 40mm]{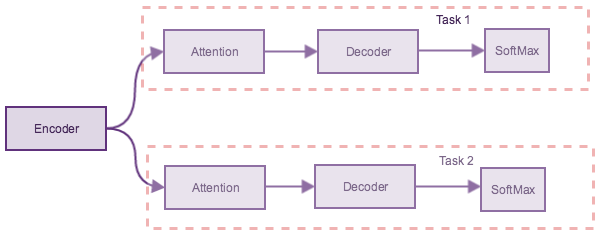}%
  \label{fig:mta}
}

\subfloat[\textit{one-to-one}: A multi-task architecture where both the encoder and decoder along with the attention layer are shared across the two tasks.]{%
  \includegraphics[clip, width = 110mm, height = 40mm]{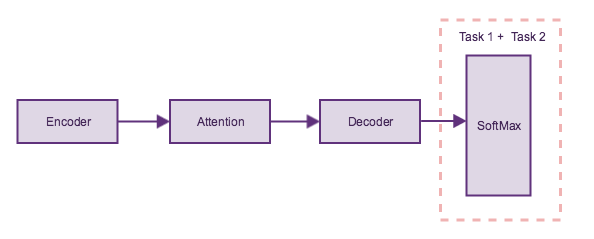}%
   \label{fig:mtb}
}
\qquad
\subfloat[\textit{one-to-shareMany}: A multi-task architecture where both the encoder and decoder along with the attention layer are shared across the two tasks, but the final softmax output layer is trained differently, one for each task.]{%
  \includegraphics[clip, width = 110mm, height = 40mm]{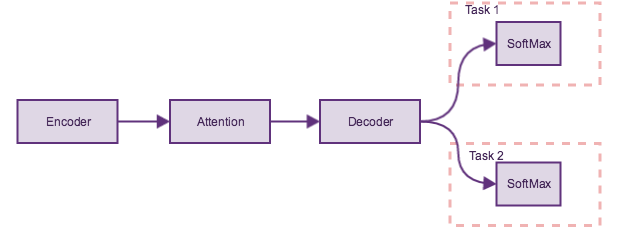}%
   \label{fig:mtc}
}
\caption{Three multi-task architectures.}
\label{figure:multi_task}
\end{figure*}

\subsection{Multi-task Setup} \label{section:multi-task}

We focus on training scenarios where multiple training sources $K$ are available. Each source $K$ can be considered a domain or a task, which consists of pairs of utterance $\vect{x}$ and annotated logical form $\vect{y}$. There are no constraints on the logical forms having the same formalism across the $K$ domains. Also, the tasks $K$ can be different, e.g., we can mix semantic parsing and syntactic parsing tasks. We also assume that given an utterance, we already know its associated source $K$ in both training and testing. 

In this work, we explore and compare three multi-task sequence-to-sequence model architectures: one-to-many, one-to-one and one-to-shareMany.  

\subsubsection{One-to-Many Architecture}

This is the simplest extension of sequence-to-sequence models to the multi-task case. The encoder is shared across all the $K$ tasks, but the decoder  and attention parameters are not shared. The shared encoder captures the English language sequence, whereas each decoder is trained to predict its own formalism. This architecture is shown in Figure~\ref{fig:mta}. For each minibatch, we uniformly sample among all training sources, choosing one source to select data exclusively from. Therefore, at each model parameter update, we only update the encoder, attention module and the decoder for the selected source, while the parameters for the other $K-1$ decoder and attention modules remain the same.

\subsubsection{One-to-One Architecture}

Figure~\ref{fig:mtb} shows the one-to-one architecture. Here we have a single sequence-to-sequence model across all the tasks, i.e., the embedding, encoder, attention, decoder and the final output layers are shared across all the $K$ tasks. In this architecture, the number of parameters is independent of the number of tasks $K$. Since there is no explicit representation of the domain/task that is being decoded, the input is augmented with an artificial token at the start to identify the task the same way as in \newcite{johnson2016google}. 

\subsubsection{One-to-ShareMany Architecture}

We show the model architecture for one-to-shareMany in Figure~\ref{fig:mtc}. The model modifies the one-to-many model by encouraging further sharing of the decoder weights. Compared with the one-to-one model, the one-to-shareMany differs in the following aspects:

\begin{itemize}
\item Each task has its own output layer. Our hypothesis is that by separating the tasks in the final layer we can still get the benefit of sharing the parameters, while fine-tuning for specific tasks in the output, resulting in better accuracy on each individual task.
\item The one-to-one requires a concatenation of all output labels from training sources. During training, every minibatch needs to be forwarded and projected to this large softmax layer. While for one-to-ShareMany, each minibatch just needs to be fed to the softmax associated with the chosen source. Therefore, the one-to-shareMany is faster to train especially in cases where the output label size is large.

\item The one-to-one architecture is susceptible to data imbalance across the multiple tasks, and typically requires data upsampling or downsampling. While for one-to-shareMany  we can alternate the minibatches amongst the $K$ sources using uniform selection.

From the perspective of neural network optimization, mixing the small training data with a large data set from the auxiliary task can be also seen as adding noise to the training process and hence be helpful for generalization and to avoid overfitting. With the auxiliary tasks, we are able to train large size modesl that can handle complex task without worrying about overfitting. 

\end{itemize}

\section{Experiments}

\subsection{Data Setup}

\begin{figure*}[h]
        \includegraphics[clip,trim=0mm 8.3cm 0 3cm, width = 150mm, height = 60mm]{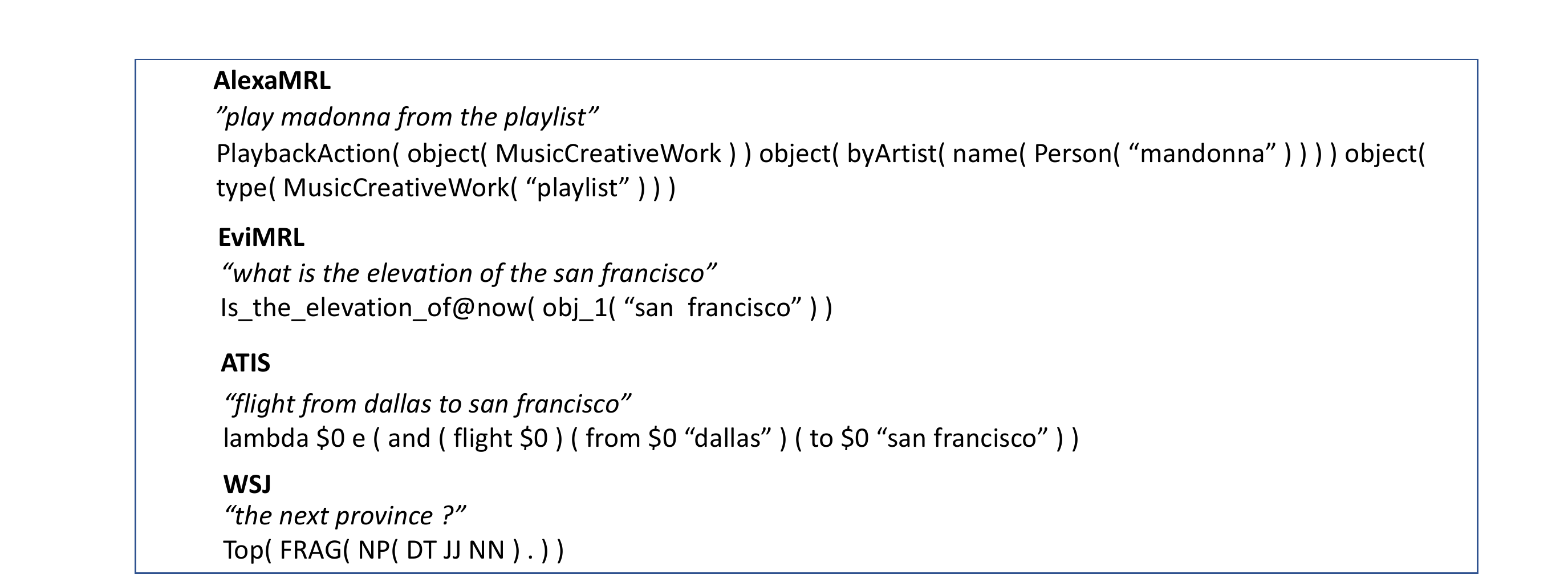}
        \caption{Example utterances for the multiple semantic formalisms}
        \label{figure:Annotation_Examples}
\end{figure*}


We mainly consider two Alexa  dependency-based semantic formalisms in use -- an Alexa meaning representation language (AlexaMRL), which is a lightweight formalism used for providing built-in functionality for developers to develop their own skills.\footnote{For details see \url{https://tinyurl.com/lnfh9py}.} The other formalism we consider is the one used by Evi,\footnote{\url{https://www.evi.com}} a question-answering system used in Alexa. Evi uses a proprietary formalism for semantic understanding; we will call this the Evi meaning representation language (EviMRL). Both these formalisms aim to represent natural language. While the EviMRL is aligned with an internal schema specific to the knowledge base (KB), the AlexaMRL is aligned with an RDF-based open-source ontology~\cite{guha2016schema}. Figure~\ref{figure:Annotation_Examples} shows two example utterances and their parses in both EviMRL and AlexaMRL formalisms.

Our training set consists of $200K$ utterances -- a fraction of our production data, annotated using AlexaMRL -- as our main task. For the EviMRL task, we have $>1M$ utterances data set for training. We use a test set of $30K$ utterances for AlexaMRL testing, and $366K$ utterances for EviMRL testing. To show the effectiveness of our proposed method, we also use the ATIS  corpora as the small task for our transfer learning framework, which has $4480$ training and $448$ test utterances~\cite{atis2007}. We also include an auxiliary task such as syntactic parsing in order to demonstrate the flexibility of the multi-task paradigm. We use $34K$ WSJ training data for syntactic constituency parsing as the large task, similar to the corpus in~\newcite{vinyals2015grammar}. 




We use Tensorflow~\cite{abadi2016tensorflow} in all our experiments, with extensions for the copy mechanism. Unless stated otherwise, we train all models for 10 epochs, with a fixed learning rate of 0.5 for the first 6 epochs and halve it subsequently for every epoch. The mini-batch size used is 128. The encoder and decoder use a 3-layer GRU with 512 hidden units. We apply dropout with probability of 0.2 during training. All models are initialized with pre-trained 300-dimension GloVe embeddings~\cite{pennington2014glove}. We also apply label embeddings with 300 dimension for the output labels that are randomly initialized and learned during training. The input sequence is reversed before sending it to the encoder~\cite{vinyals2015grammar}. We use greedy search during decoding. The output label size for EviMRL is $2K$ and for Alexa is $< 100$. For the multi-task setup, we use a vocabulary size of about $50K$, and for AlexaMRL independent task, we use a vocabulary size of about $20K$. We post-process the output of the decoder by balancing the brackets and determinizing the units of production to avoid duplicates.

\subsection{AlexaMRL Transfer Learning Experiments}

\begin{figure*}[ht]
        \includegraphics[clip,trim=0.5cm 4cm 0 2cm, width= 18cm, height=11cm]{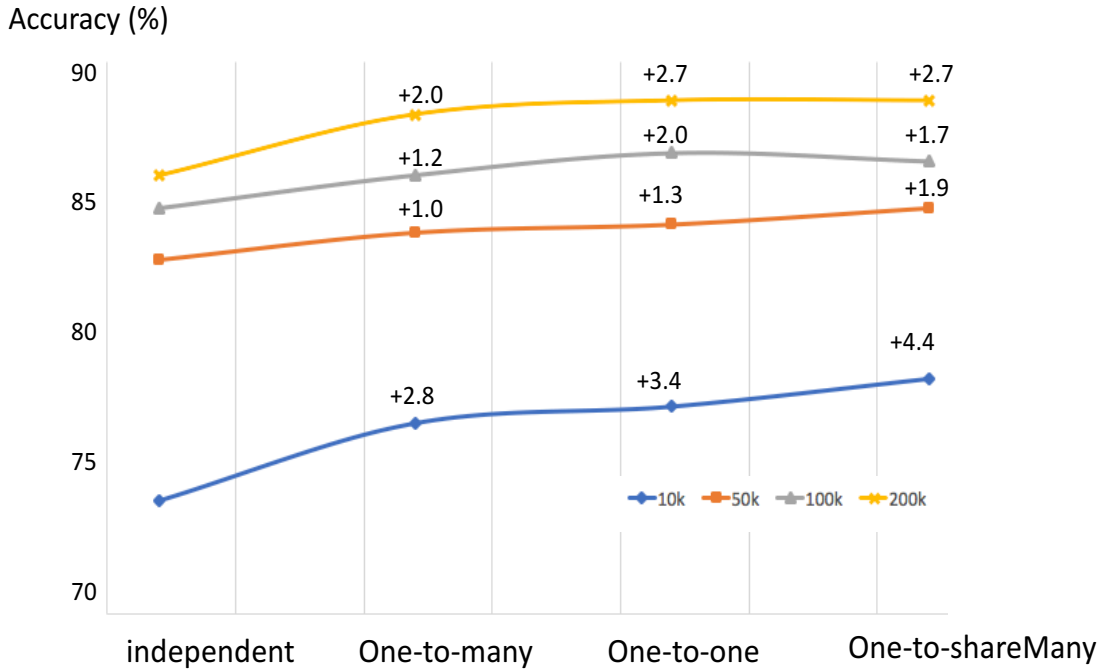}
        \caption{{\it Accuracy for AlexaMRL.}}
        \label{figure:NLU2.0_accuracy}
\end{figure*}

We first study the effectiveness of the multi-task architecture in a transfer learning setup. Here we consider EviMRL as the large source auxiliary task and the AlexaMRL as the target task we want to transfer learn. We consider various data sizes for the target task -- $10K$, $50K$ and $100K$ and $200K$ by downsampling.  For each target data size, we compare a single-task setup, trained on the target task only, with the the various multi-task setups from Section~\ref{section:multi-task} -- independent, one-to-one, one-to-many, and one-to-manyShare. Figure~\ref{figure:NLU2.0_accuracy} summarizes the results. The x-axis lists the four model architecture, and y-axis is the accuracy. The positive number above the mark of one-to-one, one-to-many and one-to-manyShare represents the absolute accuracy gain compared with the independent model. For the $10k$ independent model, we reduce the hidden layer size from 512 to 256 to optimize the performance.

In all cases, the multi-task architectures provide accuracy improvements over the independent architecture. By jointly training across the two tasks, the model is able to leverage the richer syntactic/semantic structure of the larger task (EviMRL), resulting in an improved encoding of the input utterance that is then fed to the decoder resulting in improved accuracy over the smaller task (AlexaMRL). 

We take this sharing further in the one-to-one and one-to-shareMany architecture by introducing shared decoder parameters, which forces the model to learn a common canonical representation for solving the semantic parsing task. Doing so, we see further gains across all data sizes in~\ref{figure:NLU2.0_accuracy}. For instance, in the 200k case, the absolute gain improves from $+2.0$ to $+2.7$ . As the training data size for the target task increases, we tend to see relatively less gain from model sharing. For instance, in 10k training cases, the absolute gain from the one-to-one and one-to-manyshared is $1.6$, this gain reduces to $0.7$ when we have 200k training data.

When we have a small amount of training data,  the one-to-shareMany provides better accuracy compared with one-to-one. For instance, we see $1.0$ and $0.6$ absolute gain from one-to-one to one-to-shareMany for 10k and 50k cases respectively. However, no gain is observed for 100k and 200k training cases. This confirms the hypothesis that for small amounts of data, having a dedicated output layer is helpful to guide the training.

Transfer learning works best when the source data is large, thereby allowing the smaller task to leverage the rich representation of the larger task. However, as the training data size increases, the accuracy gains from the shared architectures become smaller -- the largest gain of $4.4\%$ absolute is observed in the $10K$ setting, but as the data increases to $200K$ the improvements are almost halved to about $2.7\%$.

In Table \ref{table:model_size}, we summarize the numbers of parameters in each of the four model architectures and their step time.\footnote{In our experiment, it is the training time for a 128 size minibatches update on Nvidia Tesla K80 GPU}  As expected, we see comparable training time for one-to-many and one-to-shareMany, but 10\% step time increase for one-to-one. We also see that one-to-one and one-to-shareMany have similar number of  parameter, which is about 15\% smaller than one-to-many due to the sharing of weights. The one-to-shareMany architecture is able to get the increased sharing while still maintaining reasonable training speed per step-size.


\begin{table}[]
\centering
\begin{tabular}{@{}lllll@{}}
\bf  Model architecture & \bf param. size & \bf step time  \\\midrule
independent                & 15 million        & 0.51       \\
one-to-many         & 33 million       & 0.66     \\
one-to-one               & 28 million        & 0.71                \\
one-to-shareMany               & 28 million        & 0.65           
\end{tabular}
\caption{parameter size and training time comparision for independent and multi-task models}
\label{table:model_size}
\end{table}

We also test the accuracy of EviMRL with the transfer learning framework. To our surprise, the EviMRL  task also benefits from the AlexMRL task. We observe an  absolute increase of accuracy of $1.3\%$ over the EviMRL baseline.\footnote{The baseline is at $90.9\%$ accuracy for the single task sequence-to-sequence model}  This observation reinforces the hypothesis that combining data from different semantic formalisms helps the generalization of the model by capturing common sub-structures involved in solving semantic parsing tasks across multiple formalisms.

\subsection{Transfer Learning Experiments on ATIS}

Here, we apply the described transfer learning setups to the ATIS semantic parsing task \cite{atis2007}. We use a single GRU layer of 128 hidden states to train the independent model. During transfer learning, we increase the model size to two hidden layers each with 512 hidden states. We adjust the minibatch size to 20 and dropout rate to 0.2 for independent model and 0.7 for multi-task model. We post-process the model output, balancing the braces and removing duplicates in the output. The initial learning rate has been adjusted to 0.8 using the dev set. Here, we only report accuracy numbers for the independent and one-to-shareMany frameworks. Correctness is based on denotation match at utterance level. We summarize all the results in Table~\ref{table:atis_accuracy}.

\begin{table}[h]
\centering
\begin{tabular}{@{}ll@{}}
\toprule
System                        & Test accuracy \\\midrule
\multicolumn{2}{l}{\hspace{-2mm}\bf{Previous work}}      \\
\newcite{atis2007} & {\bf 84.6}          \\
\newcite{kwiatkowski2011lexical}   & 82.8          \\
\newcite{poon2013grounded}                 & 83.5          \\
\newcite{zhao2014type}         & 84.2          \\
\newcite{jia2016data}           & 83.3          \\
\newcite{dong2016language}        & 84.2          \\ \midrule
\multicolumn{2}{l}{\hspace{-2mm}\textbf{Our work}}                  \\
Independent model             & 77.2          \\
+ WSJ constituency parsing    & 79.7          \\
+ EviMRL semantic parsing     & 84.2          \\ \bottomrule
\end{tabular}
\caption{Accuracy on ATIS}
\label{table:atis_accuracy}
\end{table}

Our independent model has an accuracy of $77.2\%$, which is comparable to the published baseline of $76.3\%$ reported in~\newcite{jia2016data} before their data recombination. To start with, we first consider using a related but complementary task -- syntactic constituency parsing, to help improve the semantic parsing task. By adding WSJ constituency parsing as an auxiliary task for ATIS, we see a $3\%$ relative improvement in accuracy over the independent task baseline. This demonstrates that the multi-task architecture is quite general and is not constrained to using semantic parsing as the auxiliary task. This is important as it opens up the possibility of using significantly larger training data on tasks where acquiring labels is relatively easy.

We then add the EviMRL data of $>1M$ instances to the multi-task setup as a third task, and we see further relative improvement of $5\%$, which is comparable to the published state of the art~\cite{atis2007} and matches the neural network setup in~\newcite{dong2016language}.

\section{Conclusion}

We presented sequence-to-sequence architectures for transfer learning applied to semantic parsing. We explored multiple architectures for multi-task decoding and found that increased parameter sharing results in improved performance especially when the target task data has limited amounts of training data. We observed a 1.0-4.4\% absolute accuracy improvement on our internal test set with 10k-200k training data. On ATIS, we observed a $>6\%$ accuracy gain.

The results demonstrate the capabilities of sequence-to-sequence modeling to capture a canonicalized representation between tasks, particularly when the architecture uses shared parameters across all its components. Furthermore, by utilizing an auxiliary task like syntactic parsing, we can improve the performance on the target semantic parsing task, showing that the sequence-to-sequence architecture effectively leverages the common structures of syntax and semantics. In future work, we want to use this architecture to build models in an incremental manner where the number of sub-tasks $K$ continually grows. We also want to explore auxiliary tasks across multiple languages so we can train multilingual semantic parsers simultaneously, and use transfer learning to combat labeled data sparsity.

\bibliography{seq2seq}
\bibliographystyle{acl_natbib}

\end{document}